\documentclass[10pt,twocolumn,letterpaper]{article}

\usepackage[pagenumbers]{cvpr}
\newcommand{\real}[1]{\mathbb{R}^{#1}}
\usepackage{bm}
\usepackage{float}
\usepackage{graphicx}
\usepackage{multicol}

\usepackage[symbol]{footmisc}
\definecolor{cvprblue}{rgb}{0.21,0.49,0.74}
\usepackage[pagebackref,breaklinks,colorlinks,allcolors=cvprblue]{hyperref}

\def\paperID{16345} 
\def\confName{CVPR}
\def\confYear{2025}

\title{Abnormality-Driven Representation Learning for Radiology Imaging}

\author{Marta Ligero$^1$\footnote[1]{}~, \and Tim Lenz$^1$\footnote[1]{}~, \and Georg Wölflein$^1$, \and Omar S.M. El Nahhas$^1$, \and Daniel Truhn$^2$,  ~~~~~~~~Jakob Nikolas Kather$^1$ \\
\normalsize $^1$Else Kroener Fresenius Center for Digital Health, Medical Faculty Carl Gustav, $^2$RWTH Aachen University \\
{\tt\small \{tim.lenz,marta.ligero\_hernandez,jakob\_nikolas.kather\}@tu-dresden.de}
}

\begin{document}
\maketitle
\begin{abstract}
To date, the most common approach for radiology deep learning pipelines is the use of end-to-end 3D networks based on models pre-trained on other tasks, followed by fine-tuning on the task at hand. 
In contrast, adjacent medical fields such as pathology, which focus on 2D images, have effectively adopted task-agnostic foundational models based on self-supervised learning (SSL), combined with weakly-supervised deep learning (DL).
However, the field of radiology still lacks task-agnostic representation models due to the computational and data demands of 3D imaging and the anatomical complexity inherent to radiology scans.
To address this gap, we propose \textsc{Clear}, a framework for radiology images that uses extracted embeddings from 2D slices along with attention-based aggregation for efficiently predicting clinical endpoints. As part of this framework, we introduce lesion-enhanced contrastive learning (\textsc{LeCL}), a novel approach to obtain visual representations driven by abnormalities in 2D axial slices across different locations of the CT scans.
Specifically, we trained single-domain contrastive learning approaches using three different architectures: Vision Transformers, Vision State Space Models and Gated Convolutional Neural Networks.
We evaluate our approach across three clinical tasks: tumor lesion location, lung disease detection, and patient staging, benchmarking against four state-of-the-art foundation models, including BiomedCLIP.
Our findings demonstrate that \textsc{Clear} using representations learned through \textsc{LeCL}, outperforms existing foundation models, while being substantially more compute- and data-efficient.
\end{abstract}\footnote[0]{$^*$ Equal contribution}    
 \section{Introduction}
 \label{sec:intro}
\begin{figure*}[h!]
    \centering
    \includegraphics[width=0.8\linewidth]{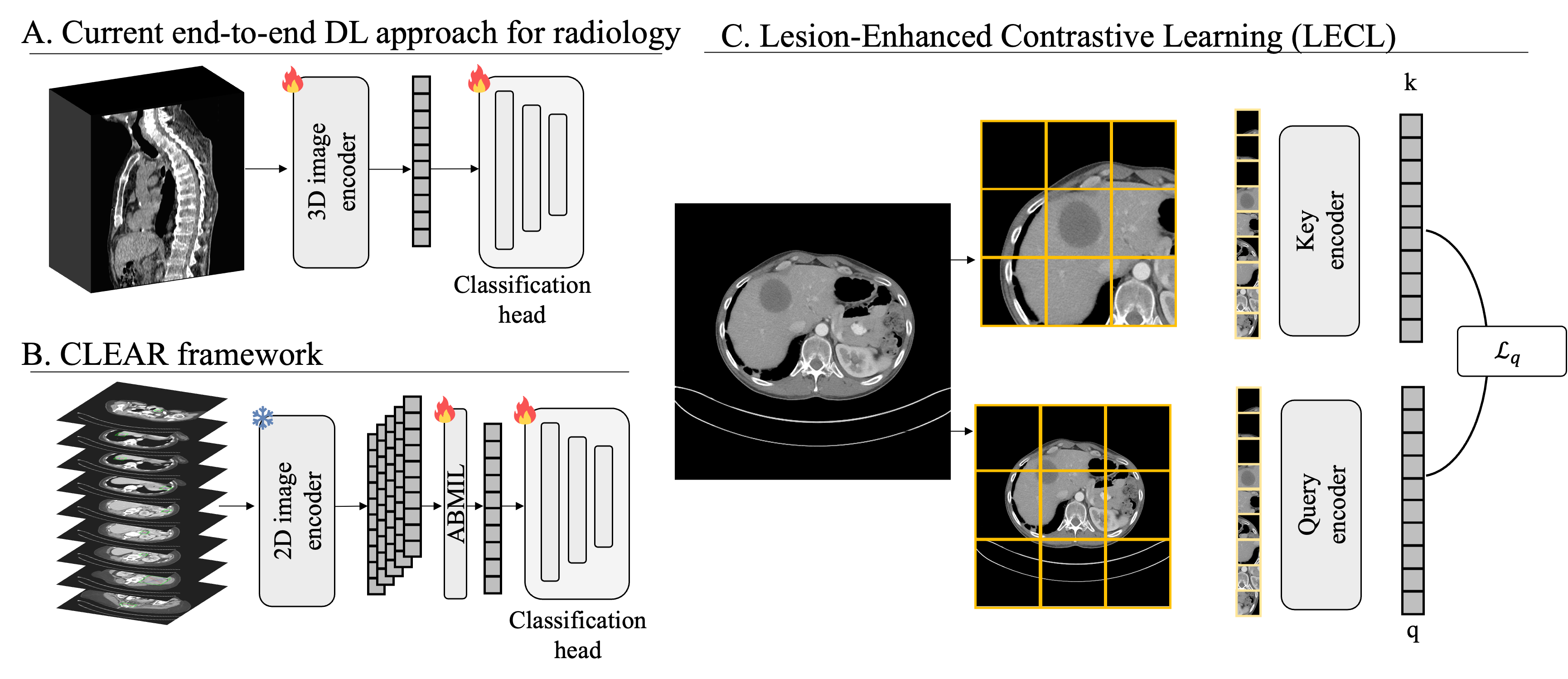}
    \caption{\textbf{Overview of the proposed framework \textsc{Clear}}.  Currently, end-to-end deep learning approaches in radiology mostly fine-tune the encoder for each specific task separately (A). We propose a weakly supervised pipeline that deploys a pretrained encoder to extract frozen embeddings, which are used for supervised training of an attention-based pooling model (B). For pretraining the feature extractor, we propose \textsc{LeCL}, a semi-supervised algorithm that guarantees that the abnormalities are within the crops of the images (C).}
    \label{fig:overview}
\end{figure*}

Recent advances in precision oncology have highlighted the need for artificial intelligence (AI) systems capable of analyzing whole-body radiology images to characterize metastatic cancer patients \cite{prelaj2024AIbiomarkers}. The development of spatial biomarkers from radiology predominantly consists of the implementation of either handcrafted features (radiomics), or end-to-end deep learning (DL) pipelines (\cref{fig:overview}A) \cite{bera2022reviewnature}. These approaches, however, both require manual or automated scan selection, followed by identification and annotation of lesions \cite{perez2024AIoncology}. With DL systems demanding extensive annotated datasets for specific tasks, these resource-intensive processing requirements represent a substantial bottleneck in current biomarker development pipelines. As a result, radiology has fallen behind other medical imaging fields such as pathology, particularly in the adoption of emerging technologies like vision transformers and state space models.

In the pathology field, there are well-established pipelines that combine the extraction of features from smaller image patches using task-agnostic foundation models combined with attention-based aggregation methods for the final downstream task prediction \cite{ElNahhas2024protocol}. This alleviates the need for tumor segmentation and extensive image preprocessing. Inspired by such methods, we investigate whether similar workflows could be adapted for radiology.  However, a critical component of such workflows is a foundation model that can extract imaging representations that generalize to different downstream tasks without the need for fine-tuning \cite{chen2024uni}, which is currently lacking in radiology. There are some domain-specific foundation models in radiology~\cite{Pai2024,blankemeier2024merlin,hamamci2024ct-clip} as well as general-purpose foundation models that encompass the entire spectrum of medical imaging, including radiology among others \cite{zhang2024biomedclip}. Nevertheless, existing approaches tend to narrowly focus on specific anatomical regions, while others overgeneralize across distinct imaging modalities.
To address these challenges, we propose Contrastive Learning-based Embeddings for Attention-based Radiology (\textsc{\textbf{Clear}}) (see \cref{fig:overview}B), a domain-specific foundation model for CT imaging using contrastive learning. We opted to develop our models to operate on stacks of two-dimensional images instead of using the 3D images directly, as this reduces computational requirements, which makes it feasible to perform self-supervised learning (SSL). However, this approach requires effective aggregation strategies for development of patient-level biomarkers. Attention-based Multiple Instance Learning (ABMIL) emerges as a promising aggregation strategy, potentially streamlining radiology workflows by automatically identifying informative slices, independent of acquisition parameters, preprocessing methods, or non-pathological regions.
Towards this purpose, our contributions are as follows: 
\begin{itemize}
    \item We introduce \textsc{Clear}, the first DL framework for radiology images that enables the development of diverse clinical applications without task-specific fine-tuning. 
    \item As part of this framework, we propose a novel semi-supervised \textbf{L}esion \textbf{e}nhanced \textbf{C}ontrastive \textbf{L}earning (\textbf{\textsc{LeCL}}) method for CT scans, and compare it with MoCo-v3~\cite{chen2021mocov3}, in terms of the quality of its feature representations for detecting abnormal lesions throughout the whole body (see \cref{fig:overview}C).
    \item We performed a comprehensive analysis of different 2D-based model architectures, including Vision Transformers (ViT), Vision State Space Models (VSSM) and gated Convolutional Neural Networks (CNN), to develop effective foundation models.
    \item A post hoc interpretability analysis that visualizes model attention across CT slices, enabling to assess whether the model focuses on regions containing pathological findings.
\end{itemize}

\section{Related work}
\label{sec:relatedwork}
\subsection{Radiology-based biomarkers for precision oncology}
In the last decade, with advancements in the field of machine learning and computer vision, several studies aimed to quantify tumor phenotypes from radiology images for the development of non-invasive biomarkers using handcrafted features, also known as radiomics~\cite{Aerts2014}. However, the implementation of radiomics features for imaging representation is restricted to specific regions of interest and to a fixed set of features. On the other hand, with the advancements in deep learning architectures and transfer learning, some studies have demonstrated the use of deep learning to predict response directly from radiology images~\cite{Jiang2023MriDL, zhou2023DLtreatmentresponse,jin2021DLpredicting,Lao2017Deeppred}. These studies are mostly trained from scratch or rely on pre-trained models designed for specific tasks. Additionally, most of these studies require large datasets and lesion delineation or bounding boxes. Initial efforts have been done in the development of foundation models for imaging biomarkers from radiology images using self-supervised deep learning models~\cite{Pai2024}. However, it is still limited to a specific region of interest, requiring manual annotation of the images. Our study aims to overcome this limitation by providing a lesion aware semi-supervised deep learning method that is trained to learn abnormalities from the data.

\subsection{Foundation models for Radiology}
There are several studies exploring self-supervised deep learning methods for radiology. However, most of these studies are focused on the development of such methods using radiographs. The availability of large open source datasets containing X-ray imaging~\cite{johnson2019mimic} and the two-dimensional resolution of such images, have facilitated a large research focusing on SSL methods in radiology. However, foundational models for 3D radiology imaging like Computerized Tomography (CT) and Magnetic Resonance Imaging (MRI) are still scarce~\cite{huang2023ssslradreview,Wolf_2024,CHEN2021covid}. To date, the largest foundation model for radiology remain proprietary, with their weights not being publicly available~\cite{yang2024medgemini}. In this study, we propose one of the first open source SSL foundation models for CT scans that do not require manual annotation and can be applied across the whole body. In contrast to other 3D SSL models~\cite{UniMiSS}, the use of 2D image representation combined with attention-based MIL makes our solution more flexible without any restrictions to number of tiles and slice thickness. 
On the other side, there are general-purpose foundation models like BiomedCLIP~\cite{zhang2024biomedclip} that are trained on 15 million figure-caption pairs, including radiology images together with histopathology and surgical resections. BiomedCLIP was trained in 2D images from Pubmed articles using contrastive language imaging pre-training (CLIP) approach. Our proposed method was developed on a substantially smaller dataset of CT scans, achieving a comparable performance to BiomedCLIP in the proposed tasks. Other foundation models available for CT images have also been developed in an image-text contrastive learning approach for specific imaging location and disease, such as CT-CLIP~\cite{hamamci2024ct-clip} for chest CT imaging and Merlin~\cite{blankemeier2024merlin} for abdominal CT imaging. However, these methods have not been developed for representation learning, showing a drop in performance and generalizability to other tasks compared to our proposed method.
\subsection{Weakly supervised deep learning in radiology}
Despite the success of attention-based weakly supervised deep learning methods in other medical imaging fields like pathology, most deep learning models for radiology use 3D end-to-end or pretrained methods~\cite{misera2024weakly}. Rather few studies have investigated the use of attention-based methods to aggregate imaging slices representations from CT and MRI images~\cite{Alp2024,Wei2024,Islam2021pulmonaryEmbo}. In this study, we provide evidence that approaches combining foundational frozen representations in combination with attention-based aggregation methods, that have already proven positive results in histopathology, can also be applicable for radiology imaging. The implementation of such approaches could accelerate the development of predictive models, reducing the need for resource-intensive tasks including scan and processing selection and disease annotation. With our lesion-aware semi-supervised DL approach, we aim to extract representations that have been trained to learn specifically from anomalies in the CT scan to enable a weakly supervised approach that requires no annotation. 
\begin{figure*}
    \centering
    \includegraphics[width=0.87\linewidth]{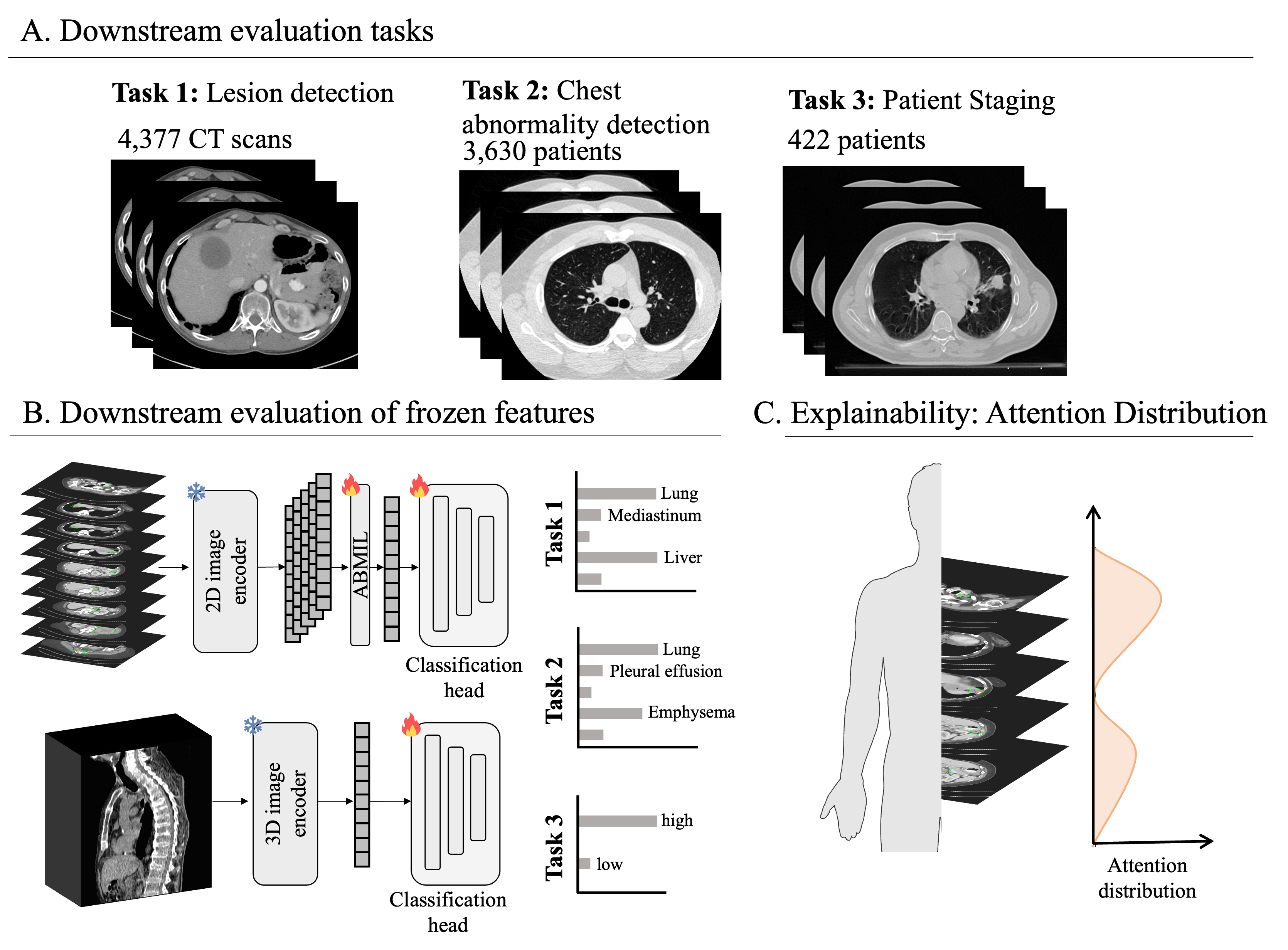}
    \caption{\textbf{Downstream task evaluation for \textsc{Clear}}. (A) We evaluated our approach in three downstream task including lesion detection (Task 1), chest abnormality classification (Task 2 and Patient staging (Task 3). (B) We compare between 2D and 3D encoders as feature extractor to evaluate the \textsc{Clear} framework for multi-task multi-label classification (Task 1 and 2) and binary classification (Task 3). (C) For explainability purposes, we explore the distribution of attention scores across CT slices }
    \label{fig:overview2}
\end{figure*}

\section{Methods}
\label{sec:methods}
In this study, we present a \textsc{LeCL}, a semi-supervised DL framework for representation learning from radiology images. Our method uses 2D axial CT scan slices and builds upon  contrastive self-supervised learning methods (\cref{sec:CL} and~\cref{sec:LECL}). We evaluate the performance of \textsc{LeCL} and Moco in different downstream tasks (\cref{sec:DWtask})

\subsection{Contrastive self-supervised learning} \label{sec:CL}
Following He et al.~\cite{He2020moco}, we interpret contrastive learning as training an encoder for a \textit{dictionary look-up task}: \\
Consider a set of encoded samples $K=\{\bm{k}_1,\bm{k}_2,\dots,\bm{k}_N\}$, which serve as the keys in a dictionary. For a given query $\bm{q}$, there exists exactly one matching key $\bm{k}^+\in K$. The contrastive loss is minimized if $\bm{q}$ is similar to $\bm{k}^+$ and dissimilar to all other keys. The InfoNCE~\cite{Oord2019} loss function is then given by: 
\begin{equation}\label{eq:moco}
\mathcal{L}_{\mathbf{q}} = -\log \frac{\psi(\mathbf{q},\mathbf{k^+})}{\sum\limits_{i=1}^N\psi(\mathbf{q},\mathbf{k}_i)}, 
\end{equation}
where $\bm{q}$ and its corresponding $\bm{k}^+$ represent feature vectors derived from different random augmentations of the same input image and $N$ is the batch size or the length of the memory queue.
The function $\psi$ is defined as follows:
\begin{equation}\label{eq:psi}
    \psi(\mathbf{x}_1,\mathbf{x}_2) = \exp(\text{sim}(\mathbf{x}_1,\mathbf{x}_2)/\tau),
\end{equation}
Here, $\tau$ represents the temperature parameter, and cosine similarity is represented by $\text{sim}(\cdot)$. To prevent feature collapse, the keys, and queries must be produced by separate encoders. Let $\theta_q$ denote the parameters of the query encoder, which includes the dense projection head. Then, the key encoder parameters $\theta_k$ are updated as follows:
\begin{equation}\label{eq:ke}
    \theta_k\leftarrow m\theta_k+(1-m)\theta_q,
\end{equation}
where \( m \in [0, 1) \) is the momentum coefficient. By using an exponential moving average of the query encoder as the key encoder, the key representations remain more stable, leading to a more stable training process.
For our experiments, we adapted the public MoCo-v3~\cite{chen2021mocov3} repository.

\subsection{Lesion enhanced contrastive learning} \label{sec:LECL}

To increase the focus on the lesions of the CTs, we propose \textit{lesion enhanced contrastive learning} (\textsc{LeCL}). It ensures that the momentum encoder receives lesion-centered image crops for the annotated slices of the deep lesion dataset, while the key encoder embeds the full image of the same slice. Additionally, we experimented with an additional term $\xi = \sum_{j=1}^L\psi(\mathbf{q},\mathbf{l_j})$, with $\{l_j\}_{j\in\{1,\dots,L\}}$ denoting the set of key slices of the training set, in the denominator of the loss function to increase the weight of the key slice encodings: 
\begin{equation}\label{eq:lecl}
\mathcal{L}^{\textsc{LeCL}}_{\mathbf{q}} = -\log \frac{\psi(\mathbf{q},\mathbf{k^+})}{\sum_{i=1}^N\psi(\mathbf{q},\mathbf{k}_i)+\lambda\cdot\xi}, 
\end{equation}
where $\lambda$ denotes a weight for the introduced term.
Thereby, the dissimilarity of the annotated slice embeddings from the rest of the training samples is increased. Thus, the separation of different lesions in the feature space by downstream models should be enhanced.

\subsection{Weakly supervised learning on frozen features} \label{sec:DWtask}
The embeddings from all $K$ axial slices of the CT scan, denoted as $H=\{\bm{h}_1,\dots,\bm{h}_K\}\in\real{K\times d}$ with a feature dimension $d$, are extracted using previously described contrastive learning methods and serve as the input of the subsequent classification module. The aggregation of the slice embeddings $H$ is defined by the following MIL pooling function $f:\real{K\times d}\to\real{d}$ given by~\cite{Ilse2018Abmil}:
\begin{equation}
    \bm{z} = f(H) = \sum\limits^K_{k=1} a_k(\bm{h}_k)\cdot\bm{h}_k,
\end{equation}
where $a_k(\bm{h}_k): \real{d}\to\mathbb{R}$ is defined as:
\begin{equation}
    a_k(\bm{h}_k) = \frac{\exp\Big(\bm{w}^{\top}\tanh\big(\bm{V}\bm{h}_k^{\top}\big)\Big)}{\sum_{i}^{K}\exp\Big(\bm{w}^{\top}\tanh\big(\bm{V}\bm{h}_{i}^{\top}\big)\Big)},
\end{equation}
where $\bm{w}\in \real{p \times 1}, \bm{V}\in\real{p \times d}$ are learnable parameters and $p$ is the attention dimension.

\section{Experiments \& results}

\label{sec:results}
\begin{table*}[h!]
\centering
\caption{\textbf{Comparison of different foundation models.} AUC performance of downstream tasks. Internal validation in Deep Lesion (Task 1). The mean over five folds is reported alongside the standard deviation as subscript.}
\label{tab:main_results}
\resizebox{\textwidth}{!}{%
\begin{tabular}{l|cccccccc|c}
\toprule
Model & Abdomen & Mediastinum & Pelvis & Bone & Soft Tissue & Kidney & Liver & Lung & Average \\
\midrule
Merlin\cite{blankemeier2024merlin} & $57.6_{2.4}$ & $54.4_{0.2}$ & $50.8_{0.6}$ & $50.0_{0.0}$ & $50.0_{0.0}$ & $50.0_{0.0}$ & $50.3_{0.1}$ & $59.5_{5.1}$ & $52.8_{2.0}$ \\
CT-CLIP~\cite{hamamci2024ct-clip} & $69.1_{0.8}$ & $60.5_{2.4}$ & $57.2_{1.7}$ & $50.0_{0.0}$ & $50.7_{0.7}$ & $49.9_{0.2}$ & $56.7_{1.0}$ & $75.7_{0.5}$ & $58.7_{1.2}$ \\
SAM2~\cite{ravi2024sam2} & $\underline{84.4}_{2.4}$ & $89.0_{2.7}$ & $86.3_{6.8}$ & $54.6_{2.8}$ & $64.2_{1.4}$ & $53.8_{3.6}$ & $77.3_{4.0}$ & $88.3_{0.5}$ & $74.7_{3.5}$ \\
BiomedCLIP~\cite{zhang2024biomedclip} & $\mathbf{85.4}_{1.0}$ & $89.3_{1.9}$ & $91.8_{2.1}$ & $53.3_{4.0}$ & $66.0_{1.7}$ & $\mathbf{65.9}_{1.7}$ & $78.6_{2.4}$ & $\underline{90.9}_{0.9}$ & $77.6_{2.2}$ \\
\hline
MambaOut-MoCo & $81.3_{1.4}$ & $88.6_{1.2}$ & $\mathbf{94.8}_{0.9}$ & $53.3_{2.8}$ & $69.5_{0.9}$ & $63.9_{2.0}$ & $79.9_{2.7}$ & $90.1_{0.4}$ & $77.7_{1.7}$ \\
MambaOut-\textsc{LeCL}-1 & $81.7_{1.6}$ & $88.7_{2.2}$ & $94.2_{0.9}$ & $54.6_{3.8}$ & $68.8_{1.5}$ & $63.1_{2.0}$ & $\underline{81.5}_{1.7}$ & $90.3_{0.8}$ & $77.9_{2.0}$ \\
MambaOut-\textsc{LeCL}-0 & $82.9_{1.3}$ & $\underline{89.4}_{1.5}$ & $\underline{94.7}_{0.5}$ & $\underline{\mathbf{55.4}}_{4.4}$ & $69.6_{2.6}$ & $\underline{64.0}_{0.5}$ & $80.1_{2.2}$ & $90.8_{0.5}$ & $\mathbf{78.4}_{2.1}$ \\
\hline
VMamba-MoCo & $82.4_{2.0}$ & $\mathbf{90.5}_{0.3}$ & $92.8_{1.4}$ & $\underline{\mathbf{55.4}}_{4.5}$ & $67.3_{1.3}$ & $63.6_{3.4}$ & $79.5_{2.8}$ & $89.9_{0.4}$ & $77.7_{2.4}$ \\
VMamba-\textsc{LeCL}-1 & $80.6_{1.2}$ & $87.5_{1.8}$ & $94.6_{0.6}$ & $53.2_{2.7}$ & $\underline{71.0}_{2.2}$ & $62.3_{2.7}$ & $81.3_{2.2}$ & $\underline{90.9}_{0.5}$ & $77.7_{1.9}$ \\
VMamba-\textsc{LeCL}-0 & $82.1_{1.0}$ & $88.1_{1.8}$ & $94.5_{1.6}$ & $54.6_{3.8}$ & $\mathbf{71.9}_{2.3}$ & $63.9_{2.5}$ & $78.9_{2.2}$ & $\mathbf{91.2}_{0.5}$ & $\underline{78.2}_{2.2}$ \\
\hline
ViT-\textsc{LeCL}-0 & $81.1_{1.2}$ & $88.5_{2.3}$ & $91.4_{1.1}$ & $50.4_{0.8}$ & $67.4_{2.3}$ & $61.0_{3.1}$ & $79.3_{1.8}$ & $89.2_{0.6}$ & $76.0_{1.8}$ \\
ViT-\textsc{LeCL}-1 & $80.0_{1.6}$ & $85.9_{1.5}$ & $92.8_{1.0}$ & $50.8_{1.0}$ & $68.4_{1.6}$ & $61.5_{2.2}$ & $80.1_{1.2}$ & $90.1_{0.2}$ & $76.2_{1.4}$ \\
ViT-ConvB & $81.0_{1.9}$ & $86.4_{1.1}$ & $93.5_{0.5}$ & $51.6_{2.0}$ & $67.2_{2.9}$ & $61.7_{1.2}$ & $\mathbf{81.6}_{1.5}$ & $89.7_{0.3}$ & $76.6_{1.6}$ \\
\bottomrule
\end{tabular}}
\end{table*}

\begin{table*}[h!]
\centering
\caption{\textbf{Comparison of different foundation models.} AUC performance of downstream tasks. External validation on RadChest (Task 2).}
\label{tab:results-radchest-auc}
\resizebox{\textwidth}{!}{%
\begin{tabular}{l|cccccccc|c}
\toprule
model & Emphysema & Bronchiectasis & Pleural Effusion & Atelectasis & Fibrosis & Opacity & Calcification & Lung Nodule & Average \\
\midrule
CT-CLIP~\cite{hamamci2024ct-clip} & $50.6_{0.5}$ & $50.0_{0.0}$ & $53.2_{1.2}$ & $50.8_{0.4}$ & $50.0_{0.0}$ & $51.6_{1.0}$ & $50.8_{0.4}$ & $51.6_{0.5}$ & $51.1_{0.6}$ \\
SAM2~\cite{ravi2024sam2} & $57.6_{1.9}$ & $50.0_{0.0}$ & $64.6_{3.7}$ & $54.6_{2.4}$ & $50.4_{0.8}$ & $58.4_{1.9}$ & $57.8_{2.1}$ & $58.0_{2.3}$ & $56.4_{2.2}$ \\
Merlin~\cite{blankemeier2024merlin} & $64.0_{1.4}$ & $52.6_{0.5}$ & $67.6_{1.9}$ & $57.0_{1.4}$ & $56.2_{1.2}$ & $59.8_{0.4}$ & $\mathbf{65.8}_{1.6}$ & $55.6_{1.0}$ & $59.8_{1.3}$ \\
BiomedCLIP~\cite{zhang2024biomedclip} & $69.8_{1.6}$ & $\underline{66.6}_{1.2}$ & $75.0_{4.8}$ & $60.8_{1.2}$ & $66.8_{1.2}$ & $60.4_{1.9}$ & $62.2_{1.7}$ & $58.8_{2.2}$ & $65.0_{2.3}$ \\
\hline
MambaOut-\textsc{LeCL}-1 & $74.0_{1.4}$ & $64.6_{1.2}$ & $80.2_{1.0}$ & $62.2_{1.2}$ & $67.6_{1.2}$ & $\mathbf{61.2}_{2.3}$ & $61.8_{2.7}$ & $58.2_{2.8}$ & $66.2_{1.9}$ \\
MambaOut-MoCo & $\underline{75.0}_{1.7}$ & $66.0_{0.6}$ & $\mathbf{81.4}_{1.4}$ & $61.4_{1.4}$ & $\underline{67.8}_{1.2}$ & $59.2_{1.2}$ & $62.4_{1.2}$ & $59.8_{1.5}$ & $\underline{66.6}_{1.3}$ \\
MambaOut-\textsc{LeCL}-0 & $\mathbf{75.6}_{1.0}$ & $\mathbf{67.6}_{2.7}$ & $79.6_{1.0}$ & $\underline{63.2}_{2.3}$ & $66.8_{1.6}$ & $\underline{60.8}_{1.2}$ & $61.8_{1.3}$ & $59.8_{1.2}$ & $\mathbf{66.9}_{1.6}$ \\
\hline
VMamba-MoCo & $72.6_{0.5}$ & $63.2_{1.3}$ & $\underline{80.6}_{1.0}$ & $62.2_{0.8}$ & $64.0_{2.4}$ & $57.2_{2.0}$ & $61.8_{1.3}$ & $\mathbf{61.8}_{1.9}$ & $65.4_{1.5}$ \\
VMamba-\textsc{LeCL}-1 & $73.6_{2.0}$ & $63.4_{1.7}$ & $78.6_{0.5}$ & $62.6_{0.5}$ & $67.6_{4.6}$ & $58.2_{1.2}$ & $62.4_{1.4}$ & $59.4_{1.6}$ & $65.7_{2.1}$ \\
VMamba-\textsc{LeCL}-0 & $73.8_{0.8}$ & $65.0_{1.1}$ & $77.0_{0.6}$ & $60.6_{1.2}$ & $\mathbf{68.2}_{1.2}$ & $59.2_{0.8}$ & $\underline{63.2}_{1.2}$ & $\underline{60.0}_{1.4}$ & $65.9_{1.1}$ \\
\hline
ViT-ConvB & $65.4_{3.6}$ & $50.4_{0.5}$ & $77.2_{1.5}$ & $60.2_{0.8}$ & $50.6_{0.5}$ & $56.6_{2.1}$ & $62.2_{1.0}$ & $57.6_{1.0}$ & $60.0_{1.7}$ \\
ViT-\textsc{LeCL}-0 & $73.0_{0.9}$ & $59.4_{3.5}$ & $76.2_{0.8}$ & $62.0_{1.8}$ & $66.2_{3.7}$ & $59.8_{1.5}$ & $62.4_{0.8}$ & $59.6_{0.8}$ & $64.8_{2.1}$ \\
ViT-\textsc{LeCL}-1 & $72.8_{1.0}$ & $63.6_{2.8}$ & $78.0_{1.8}$ & $\mathbf{64.0}_{1.4}$ & $64.8_{2.5}$ & $56.4_{2.1}$ & $62.6_{1.2}$ & $58.8_{1.3}$ & $65.1_{1.9}$ \\
\bottomrule
\end{tabular}}
\end{table*}

\subsection{Cohort description}
\paragraph{DeepLesion} We included all axial CT slices from the DeepLesion dataset~\cite{yan2018deeplesion}, which comprises 14,601 contrast enhanced CT scans from 4,427 patients with solid tumors in different regions including bone, lung, mediastinum, liver, kidney, abdomen, soft tissue and pelvis (see \cref{fig:overview2}A). Lesion bounding boxes were available for all CT scans whereas lesion labels were available for 4,177 CT scans from 1,368 patients. This dataset was used to trained representation learning methods as well as for downstream task evaluation. Labeled CT scans were used for predicting lesion location as internal downstream task evaluation (task 1). We separated a subset of 839 patients (20\%) as a test set for evaluation. All 10,224 unlabeled CT scans and 3,538 labeled CT scans (80\%) were included for model pre-training. The downstream task was trained on the overlapping labeled CT scans and evaluated on the held out test set. 
\paragraph{RadChest} We included all axial CT slices from the RadChest dataset~\cite{draelos2021radchest}, which comprises 3,630 non-contrast enhanced chest CT scans from patients with several abnormalities in the lung including emphysema, bronchiectasis, pleural effusion, consolidation, calcification, bronchial wall thickening, atelectasis, fibrosis, opacity and lung nodules (see \cref{fig:overview2}A). This dataset was used to evaluate the performance of pre-trained models for predicting lung abnormalities as a downstream task (task 2). 
\paragraph{NSCLC-Radiomics} We included all available axial CT scans from the NSCLC-Radiomics dataset~\cite{aerts2014LUNG1}, which consists of 447 non-contrast enhanced CT scans from 422 patients with Non-Small Cell Lung Cancer (NSCLC) tumors. This dataset was used to evaluate the performance of pre-trained models for predicting patient stage defined as low (Stage I and II) and high (Stage III and IV) as a downstream task (task 3)(see \cref{fig:overview2}A). 

\begin{figure}[h!]
    \centering
    \includegraphics[width=0.9\linewidth]{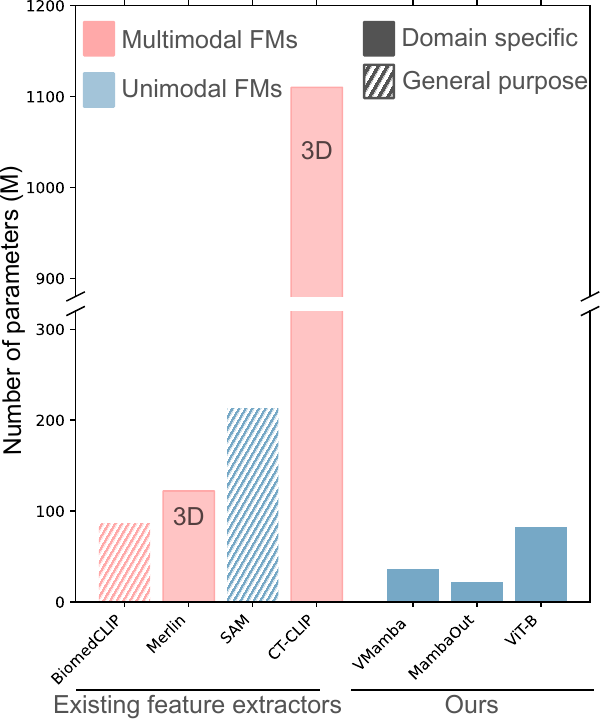}
    \caption{\textbf{Model architectures characteristics}: Number of parameters and model classification based on Multi-modal vs. Unimodal Foundation models and Domain-specific vs General-purpose (see \cref{fig:params}). We reported characteristics for the different existing foundation models (BiomedCLIP, Merlin, SAM, CT-CLIP) and the architectures used for training contrastive learning (VMamba, MambaOut, ViT).} 
    \label{fig:params}
\end{figure}
\subsection{Image preprocessing}
We considered each CT scan as a set of axial slices (up to 700 slices per patient) typically of size $512\times512~\text{px}$. Values from CT scans are stored in Hounsfield units (HU) ranging from -1024 to 1024. Different anatomical regions require specific intensity ranges to enhance structural boundaries. For pre-training, we clip the images to ranges that optimize lesion visibility, following Deep Lesion specifications (often ranging from -175 to 275 HU for abdominal window and -1500 to 500 HU for lung window). In Task 1, we implement both windows to accommodate lesions across different anatomical locations in the same patient. For Tasks 2 and 3, which focus on lung abnormalities, we apply only the lung window.

\subsection{Experimental setup}
\paragraph{Pre-training details}
We evaluate both pre-training approaches (MoCo and \textsc{LeCL}) across three different architectures (vision transformers (ViT), state space models (VMamba) and Gated CNNs (MambaOut)), assessing their suitability for radiology images (see \cref{fig:params} for architecture details). The \textsc{LeCL} approach was trained with $\lambda\in\{0,1,3,5\}$ (see \cref{eq:lecl}). 
We trained on 873,849 axial CT slices from the DeepLesion dataset. The pretraining for each model took less than 2 days (ViT-B and VMamba: 33h, MambaOut: 25h) on 4 Nvidia A100 GPUs to train for 100 epochs with a batch size of 2048 for VMamba and MambaOut and 1024 for ViT-B with a convolutional base~\cite{chen2021mocov3}. 

We used a learning rate of $1\mathrm{e}{-4}$ with 10 warm up epochs. All other parameters were configured as in the official MoCo-v3 repository~\cite{chen2021mocov3}.

\begin{figure*}[h!]
    \centering
    \includegraphics[width=0.9\linewidth]{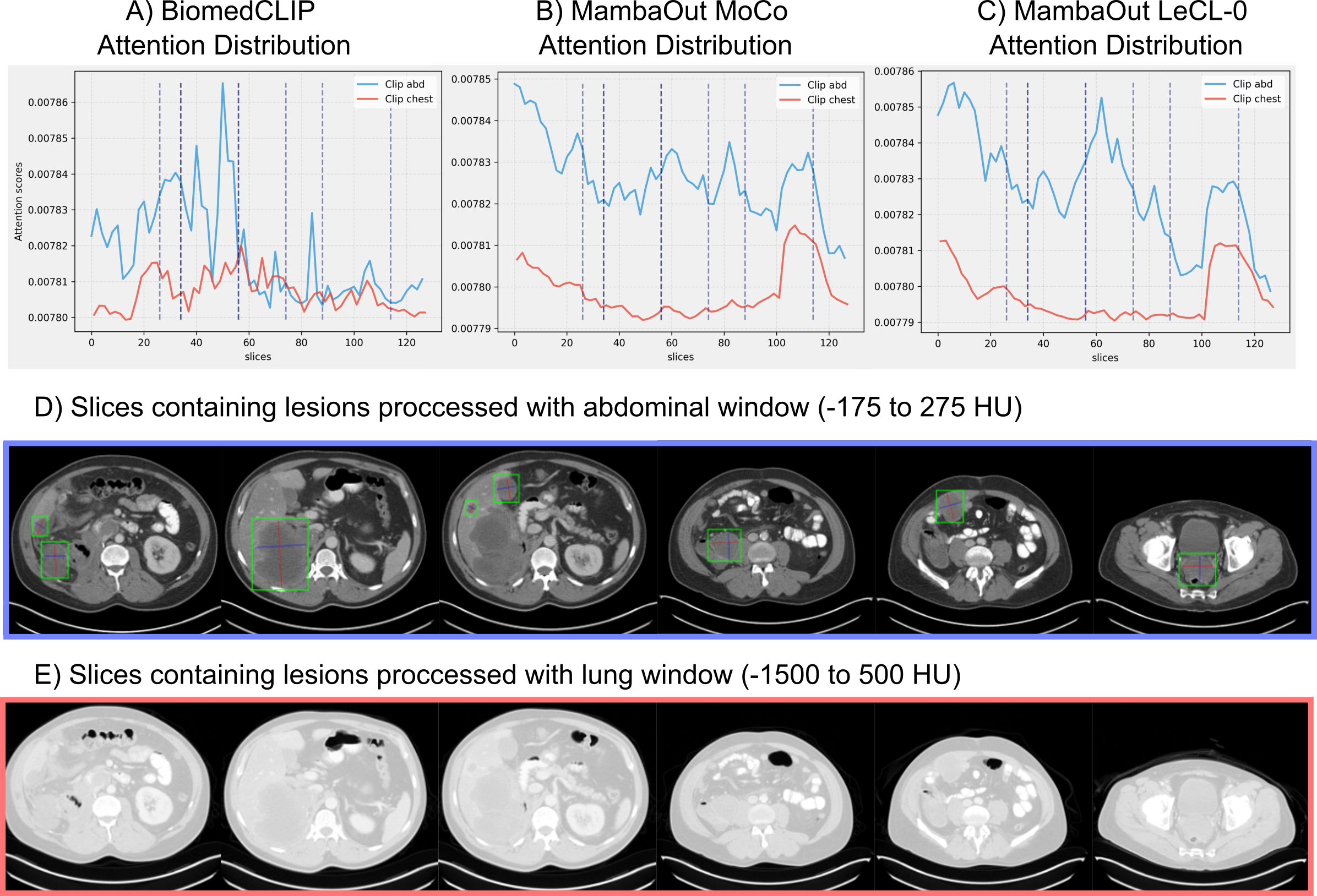}
    \caption{\textbf{Attention distribution across different slices}: We evaluated the attention distribution across slices for a patient with liver and soft tissue lesions for BiomedCLIP (A), MambaOut architecture trained using MoCo (B) and MambaOut architecture using \textsc{Lecl} approach for $\lambda$ = 0 (C). Blue represents the attention for the slices processed in abdominal window images (D) and red represents the slices processed in lung window (E). All models show higher attention to the abdominal window where the lesion is better depicted.}
    \label{fig:attention}
\end{figure*}
\begin{table}
\centering
\caption{\textbf{Comparison of different foundation models.} Performance of downstream tasks. External validation on NSCLC-Radiomics (Task 3).}
\label{tab:results-lung1}
\begin{tabular}{lllll}
\toprule
model & AUC & AUPRC & F1 \\
\midrule
CT-CLIP~\cite{hamamci2024ct-clip} & $51.6_{2.8}$ & $69.3_{0.8}$ & $63.8_{22.3}$ \\
Merlin~\cite{blankemeier2024merlin} & $60.9_{5.7}$ & $74.1_{3.4}$ & $67.2_{5.3}$ \\
SAM2~\cite{ravi2024sam2} & $61.4_{5.1}$ & $74.3_{2.8}$ & $69.8_{5.5}$ \\
BiomedCLIP~\cite{zhang2024biomedclip} & $65.2_{4.2}$ & $76.5_{2.6}$ & $67.2_{5.4}$ \\
\hline
MambaOut-\textsc{LeCL}-1 & $64.9_{5.3}$ & $76.3_{2.7}$ & $70.7_{4.8}$ \\
MambaOut-MoCo & $66.7_{4.1}$ & $77.3_{2.3}$ & $70.8_{3.7}$ \\
MambaOut-\textsc{LeCL}-0 & $\mathbf{68.3}_{5.0}$ & $\mathbf{78.3}_{2.7}$ & $\mathbf{72.3}_{5.3}$ \\
\hline
VMamba-\textsc{LeCL}-1 & $64.4_{5.7}$ & $76.1_{3.4}$ & $69.2_{4.1}$ \\
VMamba-\textsc{LeCL}-0 & $65.3_{6.3}$ & $76.7_{3.5}$ & $69.9_{3.7}$ \\
VMamba-MoCo & $65.5_{9.0}$ & $77.0_{5.1}$ & $70.4_{5.9}$ \\
\hline
ViT-\textsc{LeCL}-0 & $61.5_{7.6}$ & $74.7_{4.0}$ & $66.6_{5.0}$ \\
ViT-\textsc{LeCL}-1 & $64.8_{4.9}$ & $76.2_{2.7}$ & $69.8_{4.2}$ \\
ViT-ConvB & $\underline{67.5}_{3.7}$ & $\underline{77.8}_{2.0}$ & $\underline{71.5}_{5.9}$ \\
\bottomrule
\end{tabular}
\end{table}

\paragraph{Downstream evaluation}
We adopted the conventional linear protocol used in self-supervised learning (SSL). This approach involves freezing the backbone network weights while training only the subsequent adapter. We evaluated the performance of the frozen embeddings in three different downstream tasks and compare them against four different pre-training models as feature extractors, including both 2D approaches (BiomedCLIP~\cite{zhang2024biomedclip} and SAM2~\cite{ravi2024sam2}) and 3D approaches (CT-CLIP~\cite{hamamci2024ct-clip} and Merlin~\cite{blankemeier2024merlin}) (see \cref{fig:params} for model details). The embeddings were extracted from each axial slice from the CT scan. For 2D approaches, the adapter corresponds to an ABMIL layer~\cite{Ilse2018Abmil} followed by an MLP as classification head. In the case of 3D approaches, the adapter corresponds to the MLP as classification head (see \cref{fig:overview2}B). 

For lesion detection (Task 1) and chest disease classification (Task 2), we performed a multi-class multi-label classification approach using a binary cross entropy loss to allow for co-ocurrence of target labels. For patient staging (Task 3) we implemented a classification head with a cross-entropy loss. All models were trained using a learning rate of $1\mathrm{e}{-4}$ and batch size of 128 for 32 epochs, with early stopping after 8 epochs without improvement. We employed different validation strategies based on dataset size. For large datasets ($\geq$1,000 cases) in Tasks 1 and 2, we first separated a held-out test set and then performed 5-fold cross-validation on the remaining data for model training. The final performance was assessed by evaluating each fold's model on the common test set. For smaller datasets (Task 3), we implemented a nested 5-fold cross-validation approach, where each fold sequentially served as a test set while the remaining data was split into training and validation sets. We reported different evaluation metrics, including Area Under the Receiver Operating Characteristic (AUC), F1 score and Area Under the Precision-Recall Curve (AUPRC). 

\subsection{Representation learning for downstream radiology tasks}
Our findings show that representation learning approaches trained in vision-only settings on smaller datasets can be comparable in performance to larger architectures using multimodal approaches such as BiomedCLIP. \Cref{tab:main_results,tab:results-radchest-auc,tab:results-lung1} present the performance for our pre-trained models, using both contrastive approaches (MoCo and \textsc{LeCL}), compared to the baselines. We report the AUC here and provide the F1 scores and AUPRC in the supplementary material. 
Additionally, with this study, we highlight that the current status of representation learning for radiology is still limited, with current methods like Merlin and CT-CLIP showing relatively poor performance in all downstream tasks. Overall, our proposed approach, \textsc{LeCL}, reached superior performance in all three downstream tasks compared to other models and SSL methods. 
\begin{figure*}
    \centering
    \includegraphics[width=0.7\linewidth]{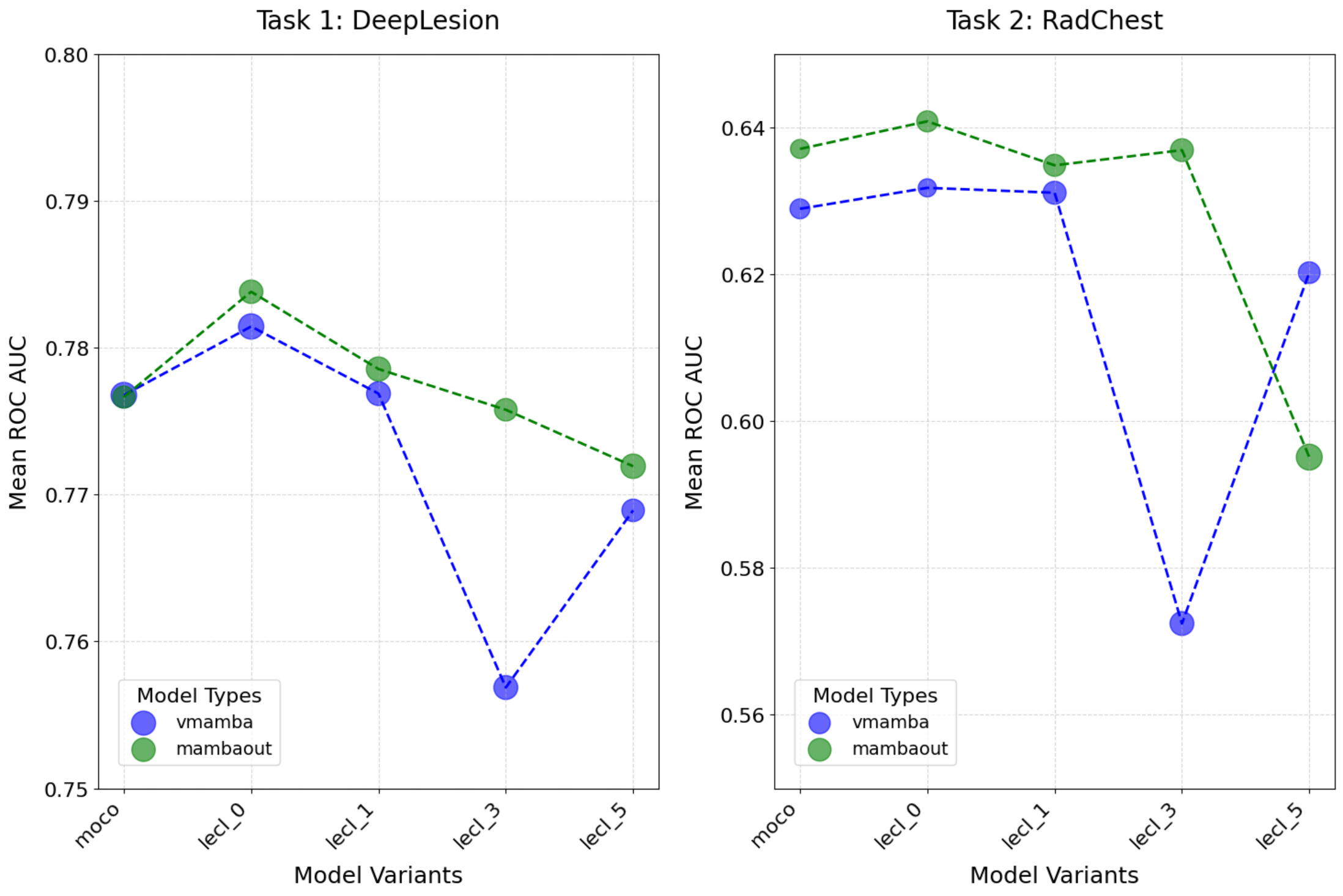}
    \caption{\textbf{Contrastive lesion weight ablation}. AUC comparison across tasks for hyperparameter $\lambda$ (see \cref{eq:lecl}).}
    \label{fig:CLablation}
\end{figure*}
\paragraph{Lesion classification} For Task 1, \textsc{LeCL-0} in combination with the MambaOut architecture reaches +0.8\% AUC compared to BiomedCLIP and +0.5\% compared to MoCo. Additionally, in more specific and anatomically complex locations like soft tissue, our \textsc{LeCL} approach outperforms BiomedCLIP by +5.90\% AUC when trained with VMamba and +3.90\% AUC when trained using MambaOut, despite being trained with substantially smaller datasets in a single-domain setting. 
\paragraph{Chest abnormality detection} For Task 2, \textsc{LeCL-0} reaches +1.90\% AUC compared to BiomedCLIP and +0.35\% compared to MoCo using the MambaOut architecture. Specifically, in tasks like detection of Emphysema, \textsc{LeCL} outperforms BiomedCLIP by +5.8\% AUC when trained with MambaOut.

\paragraph{Patient staging} For Task 3, \textsc{LeCL-0} reaches +3.10\% AUC compared to BiomedCLIP and +1.60\% AUC compared to MoCo using the MambaOut architecture. 

\subsection{Weakly supervised deep learning for CT scans}
Our findings show that our proposed framework combining representation learning and weakly supervised DL for radiology images can obtain promising results without requiring fine-tuning. By applying weakly supervised methods to the learned representations, the model is capable of selecting the most informative slices from the CT scans, ignoring images with healthy tissues or uninformative CT scan acquisitions (see \cref{fig:overview2}C). \Cref{fig:attention} shows higher attention in the tumor-containing slices for patients with different lesion types. The model effectively identifies the most relevant preprocessing methods while discarding those that poorly depict the lesions. These findings demonstrate \textsc{Clear}'s ability to streamline image processing by eliminating manual preprocessing selection and tumor annotation steps.     

\subsection{Ablation studies for model architectures}

Our ablation study investigates three distinct architectural frameworks: (1) ViT, which utilizes self-attention mechanisms, (2) VMamba, based on state space models, and (3) MambaOut, implementing gated CNNs. Overall, we find that for both contrastive learning approaches (MoCo and \textsc{LeCL}), architectures like MambaOut or VMamba achieve better performance than ViT. For the three downstream tasks MambaOut showed +2.4\%, +2.6\% and 6.8\% AUC than ViT for Task 1, 2 and 3 respectively. Similarly, VMamba showed an increase in performance compared to ViT (2.2\%,1.1\% and 3.8\% AUC for Task 1 to 3). These findings provide an explanation for the limited adoption of Transformer architectures in radiology-based applications, despite their remarkable success in other domains.

\subsection{Ablation studies for \textsc{LeCL} loss function}
\Cref{fig:CLablation} shows an ablation study for different lesion weighting parameters for contrastive learning. \textsc{LeCL} showed improved performance compared to MoCo in all tasks for MambaOut (+0.5\%, +0.35\% and +0.35\% AUC in Task 1 to 3) and for VMamba (+0.5\% AUC in Task 1 and 2). However, increasing the value of the parameter for weighting lesion representation showed a drop in performance for $\lambda\in\{1,3,5\}$. With a maximum drop for $\lambda=5$ in both MambaOut (up to -1.19\% and -4\% AUC for Task 1 and 2) and  VMamba (-0.79\% and -0.9\% AUC).

\section{Conclusion}
\label{sec:conclusion}
We introduce \textsc{Clear}, a novel framework for performing radiology image classification based on representation learning. Taking inspiration from the success of attention-based methods in pathology, our framework combines frozen embeddings with weakly supervised deep learning, showing improved performance while reducing the need for manual annotations. Within this framework, we propose \textsc{LeCL} as a method for learning lesion-aware representations. Our analysis reveals substantial limitations in current models for image representation, highlighting the need for more research into domain-specific models using representation learning approaches, as current methods fail to unlock the full potential of radiology image analysis. Through this work, we aim to encourage a shift in the radiology community by providing a framework which adapts representation learning methods to the field of radiology. 
{
    \small
    \bibliographystyle{ieeenat_fullname}
    \bibliography{main}
}


\clearpage

\maketitlesupplementary
\setcounter{page}{1}
\setcounter{section}{0}
\setcounter{figure}{0}
\setcounter{table}{0}
\setcounter{equation}{0}

\renewcommand{\thesection}{\Alph{section}}
\renewcommand{\tablename}{Supplemental Table}
\renewcommand{\figurename}{Supplemental Figure}
\begin{multicols}{2}
\setlength{\parskip}{0pt} 
\setlength{\baselineskip}{1em} 
\section{Detailed architecture for downstream tasks}
\label{sec:architect}
\begin{normalsize}
The downstream classification encoder consists of an encoding layer, am ABMIL block and a classification head. The encoder projects the embeddings to a 256 dimension using a Linear layer and a ReLU activation function~\cite{ElNahhas2024protocol}. The classification head consists of a layer norm~\cite{ba2016layernormalization} and an MLP that comprises one hidden layer and a SiLu activation function~\cite{jaume2024madeleine,Ilse2018Abmil}.
\end{normalsize}
\section{Detailed methodology for post-hoc explainability}
\label{sec:explainability}
\begin{normalsize}
The attention distribution plots for explainability show the attention weights obtained as output from the ABMIL block after being soft maxed to values between 0 and 1. Therefore, we ran inference on the CT scans of the patient with both preprocessing methods and extracted the attention layer from the patient-wise prediction. The plots from~\cref{fig:attention} show the attention distribution for each slice. The dotted vertical lines correspond to the location of the slices in a CT defined as \textit{key slices} in the Deep Lesion dataset. These \textit{key slices} and the ones surrounding them correspond to the slices that contain a lesion. Therefore, a high attention peak nearby the \textit{key slices} can indicate that the model is correctly paying attention to the areas of the body that are relevant for the prediction.  
\end{normalsize}
\end{multicols}
\begin{table*}[h!]
\centering
\caption{\textbf{Comparison of different foundation models.} F1 performance of downstream on DeepLesion (Task 1).}
\label{tab:deeplesion-f1}
\resizebox{\textwidth}{!}{%
\begin{tabular}{l|cccccccc|c}
\toprule
model & Liver & Soft Tissue & Bone & Pelvis & Abdomen & Kidney & Mediastinum & Lung & Average \\
\midrule
Merlin & $1.7_{0.6}$ & $0.0_{0.0}$ & $0.0_{0.0}$ & $3.7_{2.2}$ & $30.0_{7.7}$ & $0.0_{0.0}$ & $17.6_{1.2}$ & $39.4_{10.5}$ & $11.6_{4.7}$ \\
CT-CLIP & $25.6_{3.1}$ & $3.3_{2.7}$ & $0.0_{0.0}$ & $25.5_{4.6}$ & $57.8_{1.5}$ & $0.0_{0.0}$ & $36.7_{6.1}$ & $67.2_{0.7}$ & $27.0_{3.1}$ \\
SAM & $57.3_{3.1}$ & $43.0_{2.9}$ & $16.1_{9.3}$ & $75.9_{8.8}$ & $\underline{77.8}_{2.5}$ & $12.8_{10.2}$ & $80.9_{2.2}$ & $85.3_{0.5}$ & $56.1_{6.1}$ \\
BiomedCLIP & $65.1_{1.7}$ & $46.4_{3.2}$ & $11.1_{13.6}$ & $85.6_{1.6}$ & $\mathbf{79.1}_{1.0}$ & $\mathbf{41.0}_{2.6}$ & $81.6_{1.9}$ & $\mathbf{88.3}_{0.8}$ & $62.3_{5.2}$ \\
\hline
MambaOut-MoCo & $66.3_{2.7}$ & $52.1_{2.1}$ & $11.5_{9.7}$ & $\underline{89.1}_{0.5}$ & $74.9_{1.6}$ & $36.7_{2.9}$ & $81.8_{1.4}$ & $87.3_{0.6}$ & $62.5_{3.9}$ \\
MambaOut-\textsc{LeCL}-1 & $\underline{67.7}_{1.8}$ & $50.1_{2.0}$ & $15.9_{13.2}$ & $88.3_{0.6}$ & $75.6_{1.9}$ & $36.4_{3.2}$ & $81.4_{2.1}$ & $87.4_{0.8}$ & $62.9_{5.0}$ \\
MambaOut-\textsc{LeCL}-0 & $65.7_{1.7}$ & $52.1_{3.9}$ & $\underline{17.3}_{14.3}$ & $89.0_{0.5}$ & $76.8_{1.5}$ & $\underline{38.3}_{0.7}$ & $\underline{82.6}_{1.5}$ & $\underline{88.1}_{0.7}$ & $\mathbf{63.7}_{5.3}$ \\
\hline
VMamba-\textsc{LeCL}-1 & $\mathbf{69.1}_{2.6}$ & $\underline{53.4}_{2.9}$ & $11.0_{9.2}$ & $\mathbf{89.2}_{0.7}$ & $74.1_{1.5}$ & $32.9_{4.2}$ & $80.3_{1.7}$ & $\underline{88.1}_{0.5}$ & $62.3_{3.9}$ \\
VMamba-MoCo & $67.1_{3.2}$ & $48.0_{1.9}$ & $\mathbf{17.5}_{14.5}$ & $86.9_{1.1}$ & $76.3_{2.2}$ & $36.5_{5.2}$ & $\mathbf{83.6}_{0.6}$ & $86.6_{0.5}$ & $62.8_{5.7}$ \\
VMamba-\textsc{LeCL}-0 & $65.7_{2.2}$ & $\mathbf{55.7}_{3.3}$ & $15.7_{13.0}$ & $89.0_{1.3}$ & $75.8_{1.2}$ & $36.6_{3.6}$ & $80.5_{1.9}$ & $\underline{88.1}_{0.6}$ & $\underline{63.4}_{5.1}$ \\
\hline
ViT-\textsc{LeCL}-0 & $64.7_{1.0}$ & $47.8_{3.5}$ & $1.4_{2.8}$ & $85.2_{0.9}$ & $74.5_{1.4}$ & $30.2_{5.4}$ & $81.5_{2.2}$ & $84.9_{1.3}$ & $58.8_{2.7}$ \\
ViT-\textsc{LeCL}-1 & $65.1_{2.5}$ & $50.8_{2.9}$ & $3.1_{3.8}$ & $87.1_{1.3}$ & $73.3_{2.0}$ & $31.7_{3.8}$ & $77.6_{1.4}$ & $86.3_{0.6}$ & $59.4_{2.5}$ \\
ViT-ConvB & $67.3_{1.2}$ & $47.7_{4.8}$ & $5.7_{7.0}$ & $86.7_{0.6}$ & $74.3_{2.3}$ & $32.5_{1.2}$ & $79.0_{1.0}$ & $86.5_{0.4}$ & $60.0_{3.2}$ \\
\bottomrule
\end{tabular}}
\end{table*}

\begin{table*}[h!]
\centering
\caption{\textbf{Comparison of different foundation models.} AUPRC performance of downstream tasks on DeepLesion (Task 1).}
\label{tab:deeplesion-auprc}
\resizebox{\textwidth}{!}{%
\begin{tabular}{l|cccccccc|c}
\toprule
model & Liver & Soft Tissue & Bone & Pelvis & Abdomen & Kidney & Mediastinum & Lung & Average \\
\midrule
Merlin & $16.9_{0.2}$ & $13.9_{0.0}$ & $2.9_{0.0}$ & $15.5_{0.6}$ & $40.7_{2.4}$ & $8.5_{0.0}$ & $28.3_{0.2}$ & $38.9_{3.7}$ & $20.7_{1.6}$ \\
CT-CLIP & $21.0_{0.5}$ & $14.7_{0.7}$ & $2.9_{0.0}$ & $20.1_{1.3}$ & $50.7_{0.4}$ & $8.5_{0.0}$ & $32.2_{1.9}$ & $58.0_{0.5}$ & $26.0_{0.9}$ \\
SAM2& $39.5_{2.1}$ & $33.1_{1.4}$ & $\underline{11.1}_{4.5}$ & $62.2_{9.8}$ & $65.6_{2.2}$ & $12.3_{2.7}$ & $69.1_{2.3}$ & $79.4_{1.2}$ & $46.5_{4.2}$ \\
BiomedCLIP & $48.8_{1.3}$ & $34.2_{1.7}$ & $8.3_{6.8}$ & $75.5_{2.1}$ & $\mathbf{67.2}_{1.0}$ & $\mathbf{22.9}_{1.6}$ & $70.2_{2.2}$ & $\mathbf{82.8}_{0.8}$ & $51.2_{2.8}$ \\
\hline
MambaOut-MoCo & $49.8_{2.7}$ & $36.9_{2.3}$ & $7.6_{4.0}$ & $\underline{80.6}_{0.8}$ & $64.1_{1.3}$ & $20.0_{1.4}$ & $70.8_{1.7}$ & $81.2_{1.0}$ & $51.4_{2.1}$ \\
MambaOut-\textsc{LeCL}-1 & $51.0_{2.0}$ & $34.5_{1.1}$ & $\mathbf{11.8}_{7.4}$ & $79.4_{0.9}$ & $64.9_{1.8}$ & $20.5_{1.6}$ & $70.1_{2.5}$ & $81.3_{1.3}$ & $51.7_{3.1}$ \\
MambaOut-\textsc{LeCL}-0 & $48.6_{1.6}$ & $\underline{37.1}_{2.5}$ & $10.4_{6.2}$ & $80.5_{0.9}$ & $\underline{65.8}_{1.4}$ & $\underline{21.6}_{0.9}$ & $\underline{71.7}_{1.9}$ & $\underline{82.2}_{1.1}$ & $\mathbf{52.2}_{2.6}$ \\
\hline
VMamba-\textsc{LeCL}-1 & $\mathbf{53.1}_{2.8}$ & $\underline{37.1}_{2.2}$ & $6.4_{2.9}$ & $\mathbf{80.9}_{1.1}$ & $63.3_{1.1}$ & $17.9_{2.0}$ & $68.9_{1.9}$ & $\underline{82.2}_{0.9}$ & $51.2_{2.0}$ \\
VMamba-MoCo & $\underline{51.2}_{3.2}$ & $33.9_{1.2}$ & $10.9_{6.8}$ & $77.3_{1.6}$ & $65.5_{1.9}$ & $21.1_{1.7}$ & $\mathbf{72.9}_{0.9}$ & $79.8_{0.9}$ & $51.6_{2.9}$ \\
VMamba-\textsc{LeCL}-0 & $49.4_{2.0}$ & $\mathbf{39.7}_{2.8}$ & $\underline{11.1}_{6.9}$ & $80.5_{1.9}$ & $64.5_{1.0}$ & $20.0_{1.8}$ & $68.9_{2.2}$ & $81.5_{1.2}$ & $\underline{52.0}_{3.0}$ \\
\hline
ViT-\textsc{LeCL}-0 & $47.6_{0.7}$ & $33.6_{2.2}$ & $3.1_{0.4}$ & $74.7_{1.3}$ & $63.4_{1.1}$ & $16.2_{2.0}$ & $70.4_{2.3}$ & $76.3_{2.2}$ & $48.2_{1.7}$ \\
ViT-\textsc{LeCL}-1 & $47.8_{2.9}$ & $\underline{37.1}_{2.0}$ & $3.9_{1.6}$ & $77.7_{2.0}$ & $62.5_{1.7}$ & $17.1_{1.4}$ & $65.1_{1.5}$ & $78.4_{1.3}$ & $48.7_{1.9}$ \\
ViT-ConvB & $50.2_{1.4}$ & $33.8_{2.7}$ & $4.5_{1.9}$ & $76.8_{1.1}$ & $62.7_{1.8}$ & $17.8_{1.0}$ & $67.2_{1.0}$ & $79.8_{0.9}$ & $49.1_{1.6}$ \\
\bottomrule
\end{tabular}}
\end{table*}
\begin{table*}[h!]
\centering
\caption{\textbf{Comparison of different foundation models.} F1 performance of downstream tasks. External validation on RadChest (Task 2).}
\label{tab:results-radchest-f1}
\resizebox{\textwidth}{!}{%
\begin{tabular}{l|cccccccc|c}
\toprule
model & Emphysema & Bronchiectasis & Pleural Effusion & Atelectasis & Fibrosis & Opacity & Calcification & Lung Nodule & Average \\
\midrule
CT-CLIP & $4.0_{1.8}$ & $0.0_{0.0}$ & $11.6_{3.8}$ & $4.7_{0.6}$ & $0.0_{0.0}$ & $65.4_{1.7}$ & $\underline{82.5}_{0.4}$ & $86.8_{0.6}$ & $31.9_{1.6}$ \\
SAM2& $27.5_{6.2}$ & $0.0_{0.0}$ & $44.1_{7.7}$ & $17.6_{9.3}$ & $1.4_{2.8}$ & $63.6_{1.6}$ & $82.1_{0.8}$ & $\mathbf{87.9}_{0.4}$ & $40.5_{4.9}$ \\
Merlin & $45.8_{3.6}$ & $10.7_{3.0}$ & $50.3_{3.2}$ & $26.0_{3.8}$ & $23.1_{3.6}$ & $\mathbf{67.9}_{0.8}$ & $\mathbf{85.0}_{0.4}$ & $\underline{87.5}_{0.2}$ & $49.5_{2.7}$ \\
BiomedCLIP & $57.2_{2.8}$ & $\underline{47.6}_{1.9}$ & $61.8_{6.4}$ & $39.4_{3.8}$ & $47.6_{2.2}$ & $\underline{67.8}_{1.5}$ & $81.2_{0.7}$ & $86.1_{1.6}$ & $61.1_{3.1}$ \\
\hline
MambaOut-\textsc{LeCL}-1 & $64.3_{2.1}$ & $44.4_{2.8}$ & $72.2_{1.2}$ & $42.8_{3.2}$ & $\underline{49.5}_{1.9}$ & $65.4_{1.2}$ & $79.5_{1.4}$ & $83.8_{2.0}$ & $62.7_{2.1}$ \\
MambaOut-MoCo & $\underline{65.7}_{2.5}$ & $47.1_{1.1}$ & $\mathbf{73.6}_{1.4}$ & $40.8_{3.8}$ & $\underline{49.5}_{2.1}$ & $64.0_{2.2}$ & $81.9_{1.9}$ & $85.9_{1.6}$ & $\underline{63.6}_{2.2}$ \\
MambaOut-\textsc{LeCL}-0 & $\mathbf{66.7}_{1.6}$ & $\mathbf{49.8}_{4.8}$ & $70.1_{0.7}$ & $\underline{44.9}_{5.0}$ & $47.4_{3.1}$ & $65.1_{1.9}$ & $80.7_{3.2}$ & $86.4_{1.1}$ & $\mathbf{63.9}_{3.1}$ \\
\hline
VMamba-MoCo & $62.0_{1.0}$ & $40.5_{2.8}$ & $\underline{73.2}_{1.4}$ & $43.1_{2.4}$ & $42.3_{4.4}$ & $63.1_{1.1}$ & $80.1_{1.1}$ & $86.1_{0.8}$ & $61.3_{2.2}$ \\
VMamba-\textsc{LeCL}-1 & $63.5_{3.7}$ & $41.2_{4.1}$ & $68.4_{1.2}$ & $43.9_{2.2}$ & $49.0_{9.2}$ & $64.0_{1.9}$ & $80.6_{1.0}$ & $83.8_{1.4}$ & $61.8_{4.0}$ \\
VMamba-\textsc{LeCL}-0 & $63.8_{1.3}$ & $45.1_{2.5}$ & $66.4_{1.6}$ & $39.7_{2.4}$ & $\mathbf{51.9}_{2.3}$ & $63.5_{0.7}$ & $80.1_{1.7}$ & $85.2_{1.3}$ & $62.0_{1.8}$ \\
\hline
ViT-ConvB & $47.9_{7.9}$ & $1.5_{1.8}$ & $67.0_{2.0}$ & $38.1_{1.8}$ & $4.1_{3.4}$ & $62.6_{5.3}$ & $79.5_{2.1}$ & $85.3_{2.0}$ & $48.2_{3.9}$ \\
ViT-\textsc{LeCL}-0 & $62.3_{1.4}$ & $30.5_{9.1}$ & $64.8_{1.2}$ & $42.5_{3.3}$ & $46.0_{7.9}$ & $63.3_{2.1}$ & $81.2_{1.3}$ & $85.0_{1.0}$ & $59.5_{4.6}$ \\
ViT-\textsc{LeCL}-1 & $62.2_{2.0}$ & $41.5_{6.2}$ & $68.8_{2.7}$ & $\mathbf{47.1}_{2.7}$ & $43.8_{5.0}$ & $60.6_{3.0}$ & $81.1_{1.3}$ & $83.9_{1.7}$ & $61.1_{3.5}$ \\
\bottomrule
\end{tabular}}
\end{table*}
\begin{table*}[h!]
\centering
\caption{\textbf{Comparison of different foundation models.} AUPRC performance of downstream tasks. External validation on RadChest (Task 2).}
\label{tab:results-radchest-auprc}
\resizebox{\textwidth}{!}{%
\begin{tabular}{l|cccccccc|c}
\toprule
model & Emphysema & Bronchiectasis & Pleural Effusion & Atelectasis & Fibrosis & Opacity & Calcification & Lung Nodule & Average \\
\midrule
CT-CLIP & $31.1_{0.5}$ & $14.7_{0.0}$ & $24.2_{1.4}$ & $32.6_{0.3}$ & $14.7_{0.0}$ & $55.1_{0.5}$ & $71.5_{0.2}$ & $78.0_{0.1}$ & $40.2_{0.6}$ \\
SAM2& $38.7_{1.8}$ & $14.7_{0.0}$ & $36.4_{4.3}$ & $36.8_{2.5}$ & $15.0_{0.6}$ & $59.1_{1.1}$ & $74.5_{1.0}$ & $80.4_{0.8}$ & $44.5_{2.0}$ \\
Merlin & $43.5_{0.8}$ & $17.2_{0.8}$ & $39.5_{1.6}$ & $39.7_{1.7}$ & $22.3_{1.2}$ & $59.7_{0.2}$ & $\mathbf{78.4}_{0.9}$ & $79.5_{0.3}$ & $47.5_{1.1}$ \\
BiomedCLIP & $48.5_{1.1}$ & $35.0_{1.5}$ & $48.3_{4.8}$ & $42.1_{0.7}$ & $34.6_{1.2}$ & $60.1_{1.4}$ & $76.6_{0.9}$ & $80.7_{0.9}$ & $53.2_{2.0}$ \\
\hline
MambaOut-\textsc{LeCL}-1 & $57.4_{1.3}$ & $35.3_{2.3}$ & $60.3_{1.3}$ & $43.0_{0.9}$ & $37.0_{1.1}$ & $\mathbf{60.8}_{1.5}$ & $76.4_{1.4}$ & $80.6_{1.0}$ & $\underline{\mathbf{56.4}}_{1.4}$ \\
MambaOut-MoCo & $\underline{58.6}_{2.0}$ & $\underline{35.9}_{2.4}$ & $\underline{61.9}_{1.8}$ & $42.1_{1.1}$ & $36.0_{1.2}$ & $59.4_{0.8}$ & $76.5_{0.6}$ & $81.1_{0.6}$ & $\underline{\mathbf{56.4}}_{1.5}$ \\
MambaOut-\textsc{LeCL}-0 & $\mathbf{59.1}_{1.4}$ & $\mathbf{37.6}_{3.2}$ & $57.1_{0.9}$ & $\underline{44.9}_{2.2}$ & $34.4_{2.1}$ & $\underline{60.4}_{0.5}$ & $76.3_{0.7}$ & $81.1_{0.5}$ & $\underline{\mathbf{56.4}}_{1.7}$ \\
\hline
VMamba-MoCo & $54.7_{0.9}$ & $32.2_{0.9}$ & $\mathbf{62.2}_{1.5}$ & $42.9_{1.1}$ & $31.9_{2.7}$ & $58.2_{1.0}$ & $76.4_{0.7}$ & $\mathbf{81.9}_{0.7}$ & $55.0_{1.3}$ \\
VMamba-\textsc{LeCL}-1 & $55.7_{3.4}$ & $32.0_{1.9}$ & $55.4_{2.4}$ & $43.2_{0.6}$ & $\underline{37.8}_{5.9}$ & $58.9_{0.7}$ & $76.8_{0.7}$ & $80.9_{0.7}$ & $55.1_{2.7}$ \\
VMamba-\textsc{LeCL}-0 & $56.0_{1.0}$ & $35.0_{1.7}$ & $53.8_{2.5}$ & $40.9_{1.2}$ & $\mathbf{40.2}_{2.0}$ & $59.6_{0.5}$ & $\underline{77.1}_{0.7}$ & $\underline{81.2}_{0.5}$ & $55.5_{1.4}$ \\
\hline
ViT-ConvB & $45.7_{2.9}$ & $15.2_{0.6}$ & $54.5_{1.6}$ & $41.3_{0.9}$ & $15.2_{0.4}$ & $57.8_{1.1}$ & $76.6_{0.5}$ & $80.2_{0.4}$ & $48.3_{1.3}$ \\
ViT-\textsc{LeCL}-0 & $55.7_{1.5}$ & $24.3_{3.1}$ & $51.7_{1.0}$ & $43.2_{2.0}$ & $34.4_{4.5}$ & $59.7_{1.0}$ & $76.7_{0.5}$ & $81.1_{0.4}$ & $53.4_{2.2}$ \\
ViT-\textsc{LeCL}-1 & $55.6_{1.4}$ & $32.8_{3.7}$ & $56.8_{3.0}$ & $\mathbf{45.0}_{1.7}$ & $32.9_{3.4}$ & $57.6_{1.4}$ & $76.9_{0.6}$ & $80.7_{0.5}$ & $54.8_{2.3}$ \\
\bottomrule
\end{tabular}}
\end{table*}

\begin{figure*}[h!]
    \centering
    \includegraphics[width=0.9\linewidth]{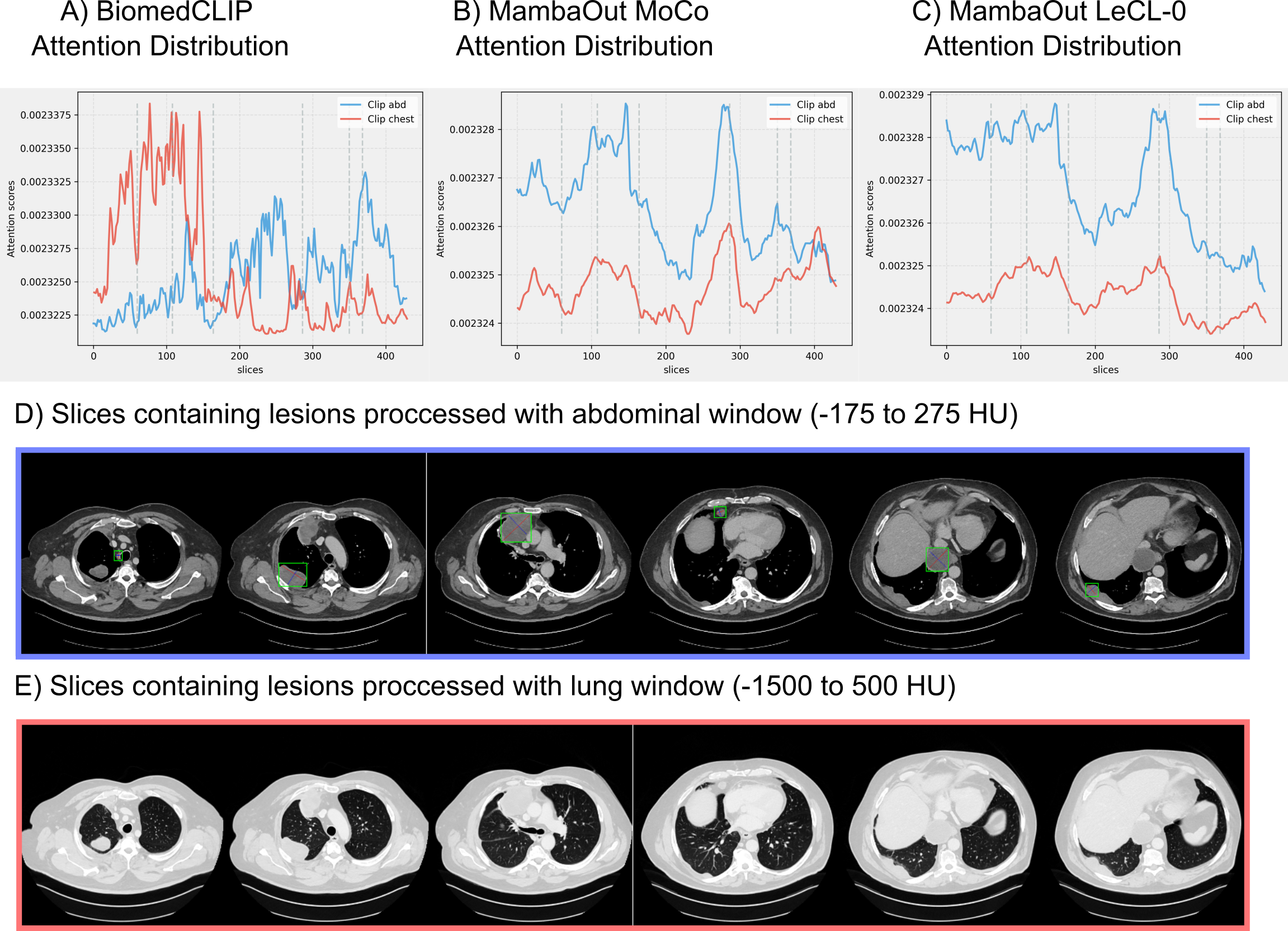}
    \caption{\textbf{Attention distribution across different slices}: We evaluated the attention distribution across slices in a patient with lung and mediastinum lesions for BiomedCLIP (A), MambaOut architecture trained using MoCo (B) and MambaOut architecture using \textsc{\textsc{LeCL}} approach for $\lambda$ = 0 (C). Blue represents the attention for the slices processed in abdominal window images (D) and red represents the slices processed in lung window (E).}
    \label{fig:attention2}
\end{figure*}

\begin{table}[h!]
    \centering
    \begin{minipage}[t]{0.48\textwidth} 
        \centering
        \caption{Overview of the number of parameters per model.}
        \label{tab:params}
        \begin{tabular}{l|c}
            \toprule
            Model & \# Parameters [M] \\
            \midrule
            BiomedCLIP~\cite{zhang2024biomedclip} & 86 \\
            Merlin~\cite{blankemeier2024merlin} & 122 \\
            SAM2~\cite{ravi2024sam2} & 213 \\
            CT-CLIP~\cite{hamamci2024ct-clip} & 1110 \\
            \hline
            Ours-VMamba & 36 \\
            Ours-MambaOut & 22 \\
            Ours-ViT-B & 82 \\
            \bottomrule
        \end{tabular}
    \end{minipage}
    \hfill
    \begin{minipage}[t]{0.48\textwidth} 
        \centering
        \caption{\textbf{Contrastive lesion weight ablation}. AUC comparison across tasks for hyperparameter $\lambda$ (see \cref{eq:lecl}).}
        \label{tab:ablation}
        \begin{tabular}{lll}
            \toprule
            Model & Task 1 & Task 2 \\
            \midrule
            VMamba-\textsc{LeCL}-5 & ${76.9}_{1.7}$ & ${61.1}_{2.0}$ \\
            VMamba-\textsc{LeCL}-3 & ${75.7}_{1.9}$ & ${56.6}_{2.3}$ \\
            VMamba-\textsc{LeCL}-1 & ${77.7}_{1.9}$ & ${62.0}_{2.1}$ \\
            VMamba-\textsc{LeCL}-0 & $\mathbf{78.1}_{2.2}$ & $\mathbf{62.0}_{1.1}$ \\
            VMamba-MoCo & ${77.7}_{2.4}$ & ${61.8}_{1.5}$ \\
            \hline
            MambaOut-\textsc{LeCL}-5 & ${77.2}_{2.1}$ & ${58.8}_{2.6}$ \\
            MambaOut-\textsc{LeCL}-3 & ${77.6}_{1.8}$ & ${62.5}_{2.1}$ \\
            MambaOut-\textsc{LeCL}-1 & ${77.9}_{2.0}$ & ${62.3}_{1.9}$ \\
            MambaOut-\textsc{LeCL}-0 & $\mathbf{78.4}_{2.1}$ & $\mathbf{62.8}_{1.7}$ \\
            MambaOut-MoCo & ${77.7}_{1.7}$ & ${62.5}_{1.3}$ \\
            \bottomrule
        \end{tabular}
    \end{minipage}
\end{table}

\end{document}